\documentclass[11pt,a4paper]{article}
\usepackage[hyperref]{acl2020}
\usepackage{times}
\usepackage{latexsym}
\usepackage{graphicx}
\usepackage{amsmath}
\usepackage{breqn}

\usepackage{microtype}

\aclfinalcopy 

\title{A Survey on Bias and Fairness in Natural Language Processing}

\author{Rajas Bansal \\
  Stanford University \\
  \texttt{rajasb@stanford.edu} \\}

\date{}

\begin{document}

\maketitle

\begin{abstract}
As NLP models become more integrated with the everyday lives of people, it becomes important to examine the social effect that the usage of these systems has.
While these models understand language and have increased accuracy on difficult downstream tasks, there is evidence that these models amplify gender, racial and cultural stereotypes and lead to a vicious cycle in many settings.
In this survey, we analyze the origins of biases, the definitions of fairness and how different subfields of NLP mitigate bias. We finally discuss how future studies can work towards eradicating pernicious biases from NLP algorithms.
\end{abstract}

\section{Introduction}
Bias is the prior information possessed by AI systems, and contrary to the emotion piqued by the word, it reflects the intelligence held by the system. Yet, this bias can be harmful, when it is the result of harmful precedents \cite{Caliskan183}. The blind application of AI systems possessing these biases runs the risk of amplifying harmful social biases which existed some 100 years ago. 

To better understand this, consider the case of retrieving web pages relevant to a query which utilizes word embeddings. Consider a simple query, like, \textit{IIT Delhi computer science student}. Now, there may be web-pages of hundreds of students which differ in nothing, except the name of the student, like John and Mary. Word embeddings attribute the field of computer science to males \cite{NIPS2016_6228}. This leads to the search engine ranking John's page above Mary's. This in turn makes women less recognized in Computer Science, thus amplifying the existing bias. 

The text corpora that is used to train State-of-the-Art algorithms themselves contain pernicious gender, racial and religious biases. \citet{garg-etal-2017-cvbed} use word embeddings as a quantitative lens through which to study historical trends - specifically trends in the gender and ethnic stereotypes in the United States. They analyze word embeddings trained over 100 years of text corpora and are able to show that their measurement of bias is highly correlated with various social movements, like the women's movement in the 1960s-1970s and Asian-American population growth in the 1960s and 1980s. 

The survey is organized as follows: Section \ref{sec:bias} outlines the sources of bias in NLP; Section \ref{sec:fairness} defines the notion of fairness; Section \ref{sec:metrics} define different metrics used to quantify bias; Section \ref{sec:debias} collates how debiasing is approached within different subfields of NLP while Section \ref{sec:current} explains how current techniques are not sufficient; Finally, Section \ref{sec:future} concludes and discussed directions for future work.

\section{Bias}
\label{sec:bias}
Data collected from humans in the real world is not homogeneous, as is usually assumed in research. Demographics of the humans labelling the data may itself influence bias the model. Data collected in the real world is heterogeneous, generated by social subgroups with their own characteristics and behaviours. A model trained on biased data may lead to unfair and inaccurate predictions. Current work in NLP focuses on curing the symptoms of these biases, without trying to drill out their origins. This leads to the developed systems fixing only a specific type of the bias (for eg. gender) and not being able to generalize to cure others (like racial biases). This paper presents the origins of the prevalent biases in order to facilitate research sensitive to these. They can be grouped into 4 major groups \cite{shah2019predictive} :-

\begin{itemize}
    \item \textbf{Label bias} Label bias emerges when the distribution of the labeled variable in the input dataset diverges substantially from the ideal distribution. Here, the label themselves are erroneous. This may be due to the lack of expertise or prejudices held by annotators.
    \item \textbf{Selection bias} Selection bias emerges due to non-representative observations – when the annotators generating the training data have a different distribution than where the model is to be applied. A famous example being the ``Wall Street Journal effect”, where syntactic parsers and part-of-speech taggers perform most accurately over language written by middle-aged white men \cite{garimella-etal-2019-womens}.
    \item \textbf{Over-amplification} This may occur during the learning process itself, where the model may pick up small difference in human factors, and amplify these to be much larger in the predicted outcomes. 
    \item \textbf{Semantic bias} Embeddings are used as an off-the-shelf source of semantic information in NLP. However, these embeddings themselves contain unintended biases, with words for certain professions like nurses, homemakers and receptionists clustering together near feminine words, and thus act as a source of semantic bias for the downstream model.
\end{itemize}

Typical word embeddings were found to associate European American names with more pleasant/positive words than African American words, and thus exhibit racial biases. While this racial bias is acknowledged by many papers, the majority of the work in this field focuses on gender bias. Other social biases like religious biases and cultural biases have not been given adequate attention and may require completely new methods to tackle them and may bring new challenges with them.

\section{Fairness}
\label{sec:fairness}
Fairness is at the heart of the problem but most prior work chooses to ignore it. The definitions have been listed so that future work realizes how different definitions of fairness may alter the problem, and which definition is most suited for their domain.

Consider the task of predicting an output variable Y given an input variable X, while remaining unbiased with respect to some variable Z. Let the predicted distribution be $\hat{Y}$. \cite{10.1145/3278721.3278779}

\begin{itemize}
    \item \textbf{Demographic Parity} A predictor $\hat{Y}$ satisfies demographic parity if $\hat{Y}$ and Z are independent. This definition says that the probability of a person being in the positive class should be independent of if the person is male or female.
    
    \item \textbf{Equality of Opportunity} A predictor $\hat{Y}$ satisfies equality of opportunity with respect to a class y if $\hat{Y}$ and Z are independent conditioned on Y = y. This says that the true positive rates should be the same for men and women.
    
    \item \textbf{Equality of Odds} A predictor $\hat{Y}$ satisfies equality of odds if $\hat{Y}$ and Z are conditionally independent given Y. This definition says that the true positive and false positive rates should be the same for men and women.
\end{itemize}

The definition of fairness to use is often a political choice, the right prefers Equality of Opportunity while the left prefers Equality of Odds.
\section{Metrics for Observing Bias}
\label{sec:metrics}
A key problem in eradication of bias is that of quantifying and measuring the bias that may be already present in the training data or model. As quantifying bias is difficult in the symbolic domain, most metrics assume that words have been projected to some vector space before measuring bias.

\begin{enumerate}
    \item \textbf{Word Embedding Assosiation Test}
    
    The WEAT test proposed by \citet{Caliskan183} is the most popular metric used to measure the bias present in word embeddings. The origins of this test are found in the Implicit Association Test (IAT) used by psychologists to show biases existing among humans using reaction time as a quantitative measure. \citet{Caliskan183} show that human biases are also reproduced in Word2Vec and GLoVe embeddings. For example, male attributes like brother, father and grandfather were found to be more correlated with science-related words than female attributes. This intuition is used to formulate the WEAT test.

    Consider two sets of target words (X,Y) (eg. programmer, engineer etc. and nurse, teacher etc.) and two sets of attribute words (A,B) (eg., he, man etc. and she, woman etc.). The null hypothesis is that there is that the attribute sets are similar to both target sets in terms of word similarity. The test measures the (un)likelihood of this hypothesis. The test statistic is :- 
    \begin{multline}
    s(X,Y,A,B) = \sum_{x \in X} s(x,A,B)\\ - \sum_{y \in Y} s(y,A,B)
    \end{multline}
    where,
    \begin{multline}
        s(w,A,B) = mean_{a \in A}cos(w,a)\\ - mean_{b \in B}cos(w,b)
    \end{multline}
    In other words, $s(w, A, B)$ measures the association of $w$ with $A$ as compared to $B$. Different variations of $A, B, X, Y$ are used in different papers.
    
    \item \textbf{Sentence Encoder Association Test}
    
    This is a simple generalization of the WEAT test for sentence encoders. Here, the target and the attribute sets contain sentences instead of words. To extend the word level test to a sentence level test, words present in the earlier sets are put into semantically bleached sentence templates such as ``This is word.'',``word is here.'', ``word are things''. \cite{may-etal-2019-measuring}
    
    \item \textbf{Projection on gender direction}
    \citet{NIPS2016_6228} define the gender bias of a word $w$ by its projection on the ``gender direction'': $\vec{w} \cdot (\vec{he} - \vec{she})$ assuming all vectors to be normalized. The higher the projection on this direction, the more bias the word is said to possess. This measurement is done, assuming that words have been partitioned into two sets gendered words (like brother, mother etc. which contain implicit gender information) and gender neutral words (eg. programmer, homemaker etc which must not contain gender information). This metric is based on the implicit assumption that there is a gender subspace which subsumes all gender information of any word.
    
    \item \textbf{Distance from gender pairs}
    Some papers use the distance from gender pairs (like he and she) as a measure of the bias in the word. Models based on this metric ensure that gender neutral words like nurse, as defined above, are equidistant from he and she.
    
    \item \textbf{Language Model Bias}
    
    \citet{bordia-bowman-2019-identifying} define the probability of a word occuring in context with gendered words as 
    \[P(w|g) = \frac{\frac{c(w,g)}{\sum_{i}c(w_i,g)}}{\frac{c(g)}{\sum_i c(w_i)}}\]
    where $c(w,g)$ is the count of the word in a specified context window and $g$ is the set of gendered words which belong to the male $(m)$ or the female $(f)$ set. The bias score of any word $w$ would then be,
    \[bias(w) = log(\frac{P(w|f)}{P(w|m)})\]
    
    \item \textbf{Gender Bias Evaluation Test-sets}
    All of the metrics discussed above quantify bias present in word embeddings, failing to quantify the bias present in the model itself. This is a very narrow view, and Gender Bias Evaluation Test-Sets (GBETs) introduced by \citet{sun-etal-2019-mitigating} provides a solution to this problem. Standard test sets in NLP fail to measure the gender bias in models. These test sets themselves contain biases (such as containing more male references than female references), so evaluation on them might not reveal biases. The goal of designing these GBETs is to provide a standardized dataset to the research community to streamline research and allow them to measure biases present in their algorithms.
    
    \textbf{Gender Swapped GBETs} The most commonly used method to construct these GBETs is gender swapping. Gender swapping is done by replacing every male definitional word with its female equivalent and vice versa. A system free from bias must give the same performance on both the male and female swapped test sets. The difference in score here acts as a proxy for the gender bias in the system.
    
    For coreference resolution, \citet{rudinger-etal-2018-gender} and \citet{zhao-etal-2018-gender} independently designed GBETs based on the gender swapping techniques and is discussed in detail later. For sentiment analysis, a GBET dataset named Equity Evaluation Corpus \cite{kiritchenko-mohammad-2017-best} is designed where every sentence in the dataset contains a gendered word along with an emotion carrying word. Gender bias is measured as the difference in emotional intensity predictions between gender-swapped datasets.
\end{enumerate}
\section{Approaches for Debiasing Models}
\label{sec:debias}
There have been numerous attempts to eradicate bias and bring fairness into NLP models for solving a variety of sub-problems like coreference resolution, semantic role labeling and machine translation. Most effort in this direction however has been in debiasing word embeddings which are used as an off-the-shelf source of semantic knowledge. Thus in the first subsection we discuss the different methods used to debias word embeddings. We then take a number of subproblems in NLP to emphasize the applications of these techniques to different domains.

Generally, methods that target bias fall into three large categories :-

\begin{enumerate}
    \item \textbf{Pre-processing} These methods acknowledge the bias that is present in the training data itself and try to alleviate this bias by transforming the input to the model itself. The downstream training algorithm is unchanged.
    
    \citet{pmlr-v97-brunet19a} identify and remove the articles from Wikipedia and the New York Times articles corpus which lead to the most increase in bias of the downstream GLoVe word embeddings. They use influence functions to approximate how the WEAT score changes due to the removal of one document from the training data. Then, the documents whose removal causes the most decrease in bias are removed. As removal of one document causes an imperceptibly small decrease in bias, a set of documents are considered for removal rather than a single document.
    
    \item \textbf{Post-processing} Sometimes, re-training the model is infeasible as it may require a large amount of resources to train the model. In such a case, post-processing methods which alter the output without considering bias at the time of training are useful. The learned model is treated as a black box in this case and the labels assigned by the model are altered in order to improve bias scores. Most methods discussed in this survey are post-processing methods.
    
    \item \textbf{In-processing} In-processing techniques try to modify and change state-of-the-art learning algorithms in order to remove discrimination during the model training process.
    
    \citet{zhao2018learning} propose a Gender Neutral Variant of GLoVe training. They point out that post-processing methods remove gender information from embeddings which may be useful for some settings such as medicine. These methods also require tagging gender neutral words which may lead to pipelining errors due to wrong identification of such words. The authors' Gender Neutral GLoVe training concentrates all socially-biased information in certain dimensions which can easily be removed if such an information is not required. Thus a word vector $w$ consists of two parts, $w = [w^{(a)};w^{(g)}]$. All gender information is concentrated into $w^{(g)}$, freeing $w^{(a)}$ from any gender influence. This is done by adding two additional loss terms to the GLoVe learning objective. One, which separates out the $w^{(g)}$ of male-definition and female-definition words far from each other (by maximizing the distance between them). The second encourages the $w^{(a)}$ of the neutral words to be in the null space of the gender direction, $v_g$. $v_g$ is estimated on the fly by averaging the difference between the male definition and the female definition words.
    
    \subsection{Word Embeddings}
    Word embeddings are a vector representation of a symbolic domain i.e words and are a fundamental component in many NLP systems. However, these contain 100 years of gender and ethnic sterotypes \cite{GargE3635}. Thus, a lot of attention has been paid to debiasing these vectors. However, \citet{Caliskan183} view this debiasing with skepticism. They view AI as perception followed by action, and they feel that debiasing alters the AI's perception and model of the world, rather than how it acts on this perception. Prejudice can anyway creep back in due to other proxies which may not be apparent \cite{gonen-goldberg-2019-lipstick}. 
    
    \subsubsection{Removing Gender Subspace}
    
    The work in this line of thinking is based on the hypothesis that all gender, racial and religious biases and information can be subsumed by a particular subspace of word embeddings. These algorithms thus try to reduce the information that the gender neutral words possess in the gender-biased subspace. Debiasing involves two steps :-
    
    \begin{enumerate}
        \item \textbf{Identify gender subspace}
        \item \textbf{Neutralize vectors}
    \end{enumerate}
    
    First all words are separated into male-definitional words ($\Omega_m$), female-definitional words ($\Omega_f$) and gender neutral words ($\Omega_n$). Many words in $\Omega_m$ and $\Omega_f$ are paired together into complements (for eg. (he,she), (brother,sister), (John,Mary) etc.). The difference in the vector of these words i.e $\vec{he} - \vec{she}$ acts as the gender direction. Instead of taking just one difference, general practice is to take all such difference and run a PCA algorithm, which yields the subspace which explains the most variation in this space of gender direction. This is done by stacking all the above differences into a matrix \textbf{C}. The bias subspace, B is then defined to be the first $k$ rows of SVD(\textbf{C}).
    
    The neutralize step removes gender bias from all the vectors, $\vec{w} \in \Omega_n$. It does this by re-embedding all the word vectors through the following process :-
    \[\vec{w} := \frac{\vec{w} - \vec{w_B}}{||\vec{w} - \vec{w_B}||}\]
    where $\vec{w_B}$ is the projection of w onto the bias subspace B. This removes the gender component from all neutral words. 
    
    
    
    \subsubsection{Data Augmentation}
    Sometimes the training dataset may have a disproportionate number of references to one gender. To mitigate this, \citet{zhao-etal-2018-gender} propose gender swapping in the training dataset which equalizes the references to each gender in this dataset. This is different from a GBET as GBETs were used to measure the bias that may be present in the model while this approach debias predictions by training on a gender-balanced dataset. 
    When applied to coreference resolution model originally trained on OntoNotes 5.0 which was tested on WinoBias, gender augmentation lowered the difference between F1 scores on pro- and anti-stereotypical test sets significantly, which indicates the model does not make predictions motivated by gender.
    
    Complementary to augmentation, \citet{pmlr-v97-brunet19a} propose removing those documents which contain the most amount of bias. They show that removing the 10 most influential (according to their metric) documents from the training set led to a 40\% decrease in the WEAT effect size.
    
    \subsection{Sentence Encoding}
    
    Recently, opposed to word embeddings, sentence representations thorugh ELMo, BERT and GPT have become the prefered choice for encoding text. As these models contain contextual word embeddings in place of a single embedding for each word, traditional WEAT tests cannot be used here. Bias is measured in such cases using the SEAT test \cite{Liang2019TowardsDS,may-etal-2019-measuring}. The SEAT test is a very crude generalization of the WEAT test as discuessed. 
    \citet{may-etal-2019-measuring} show that the biases that were shown by \citet{Caliskan183} are reproduced for BERT and ELMo. The Hard Debias method estimates the bias subspace by computing the PCA of gendered sentence templates followed by removal of the component of the sentence encoding in the bias subspace.
    
    While many papers show that these debiasing methods reduce SEAT scores, they also report that the SEAT does not seem to be an adequate metric. Certain counter-intuitive trends are observed in the bias present in sentence embeddings as compared to the word embeddings. This may be due to two reasons, one, the templates formed are not as semantically bleached as expected and two, cosine similarity is not a good metric to measure sentence similarity as training for BERT and ELMo has been done on different objectives. Thus a better measure of bias in sentence representations is required.
    
    \citet{Kurita_Vyas_Pareek_Black_Tsvetkov_2019} propose a new measure of bias for sentence representations which shows better correlation with human biases than SEAT and addresses some of the problems mentioned above. The process for measuring bias by correlating a [TARGET] (gendered word, in this case, he) to an [ATTRIBUTE] (in this case programmer) is as follows :- 
    \begin{enumerate}
        \item Prepare a template sentence e.g. ``[TARGET] is a [ATTRIBUTE]''
        \item Replace [TARGET] with [MASK] and compute 
        \[p_{tgt}= P([MASK]=[TARGET])\]
        \item Replace both [TARGET] and [ATTRIBUTE] with [MASK], and compute prior probability 
        \[p_{prior} = P([MASK]=[TARGET])\]
        \item Compute the association between target and attribute as $log(\frac{p_{tgt}}{p_{prior}})$
    \end{enumerate}
    The bias can be calculated by subtracting the association scores for he/she.
    
    \subsection{Coreference Resolution}
    Coreference resolution is the task aimed at identifying phrases (mentions) refering to the same entity. \citet{zhao-etal-2018-gender} created a novel challenge corpus, WinoBias which follows the winograd format and contains references to people using a vocabulary of 40 occupations. Figure 1 shows the 2 types of sentences that are present in the WinoBias corpus.
    
    \begin{figure}
      \includegraphics[width=225pt]{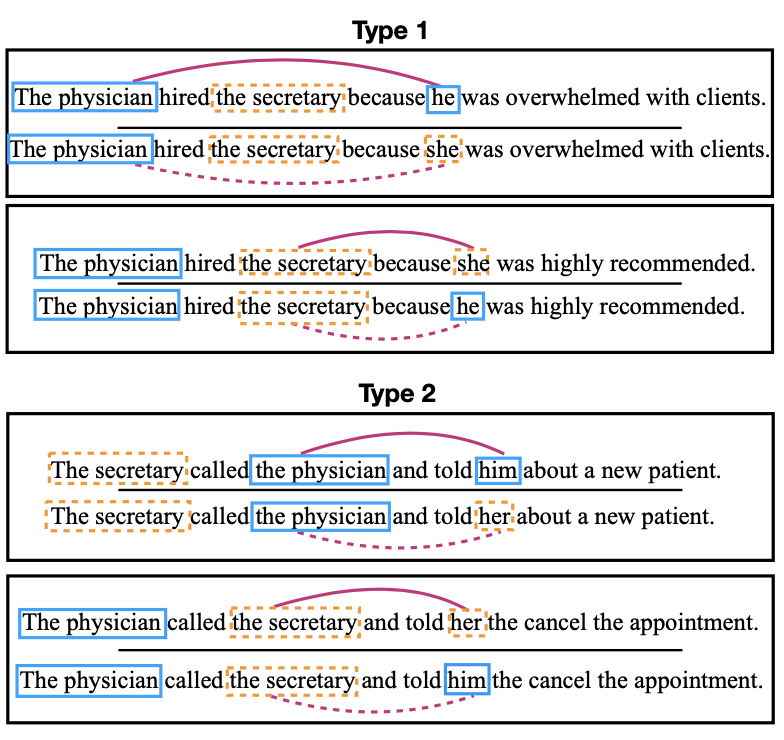}
      \caption{\cite{zhao-etal-2018-gender} The WinoBias data with gender swapping. Male and female entities are marked in blue and orange respectively. The pro- and anti-stereotypical are shown through solid and dashed lines respectively. The gender should be irrelevant for the coreference system.}
    \end{figure}
    
    Typical systems were found to give poor accuracy on the WinoBias dataset. This is because the widely used OntoNotes 5.0 dataset uses male gendered pronouns to refer to 80\% entities. In order to reduce this bias, an additional training corpus was created by swapping all male entities for female entities and vice versa. The SoTA end-to-end system was then trained on the original augmented with the swapped corpus. Names were anonymized to remove the bias that a gendered name may carry with itself. The authors were able to show that the SoTA algorithms performed poorly on the WinoBias dataset. The anonymization, usage of bias free embeddings and the augmentation of data, as discussed were enough to improve the performance of these systems on the dataset. As the WinoBias evaluation may be out of domain for a system trained on OntoNotes, the authors also compared the difference in accuracy of all systems on the OntoNotes dev set and its gender swapped version. The debiased system did not show a significant difference in accuracy on swapping the gender.

    \subsection{Machine Translation}
    The example most widely used to show bias in machine translation is the English sentence "She works in a hospital, my friend is a nurse". Here my friend is translated to ``amiga'' (girl friend) in Spanish, while the sentence ``She works in a hospital, my friend is a doctor'' translates my friend to amigo (boy friend) in Spanish. \citet{escude-font-costa-jussa-2019-equalizing} carry out a study of biases in MT from English to Spanish. In addition to the corpus available from WMT, they have created an additional test set to study gender bias. Motivated from the above example the test set is built using a sentence pattern ``I've known \{her, him, proper-noun\} for a long time, my friend works as a \{occupation\}.'' They used debiased word embeddings constructed through methods mentioned above and were able to show a huge improvement on the custom test set over the SoTA baseline. In fact, they were able to show improvement of 1 BLEU point over the original test set showing that the performance does not degrade when one uses these debiased embeddings.
    
    Human translators rely on contextual information to infer certain aspects like gender of the speaker in order to make the best translation. In absence of such context, like the identity of the speaker, NMT make the statistically most likely choice which may be gender biased. \citet{vanmassenhove-etal-2018-getting} thus give more emphasis on speaker information being present in parallel data. They have thus released a large corpus of parallel data tagged with speaker information (name, gender, age, date of birth, euroID and date of the session). They then trained a NMT system which was given the gender of the speaker as the first token. The authors were able to show a 1 point improvement in BLEU score with the most improvement seen in sentences having female speakers.
    
    \subsection{Semantic Role Labeling}
    \citet{menshopping} consider the task of visual semantic role labeling (vSRL) where the goal is to predict activities and the roles that objects play in it. Biases exist in such a setting too. In the imSitu training set, only 33\% of cooking images have man as the agent role. After training a CRF, this bias is amplified: less than 16\% of cooking images are labeled with man as the agent. They use a novel constrained inference framework called RBA (Reduced Bias Amplification). Here the Language Model bias formulation (4.5) is used to measure bias of the system. Bias amplification is measured by comparing the bias on the training set to that on the test set.
    
    In order to reduce bias amplification, the authors inject constraints enforcing that the test bias should remain in certain limits of the training bias. These constraints need to be applied at the corpus level and thus would require solving a large inference problem. Therefore, an approximate inference algorithm based on Lagrangian relaxation is used. Let $b^{star}$ be the gender ratio present in the training set. Let $y^{i}_{M}$ denote the $i^{th}$ test instance being assigned the agent as man. Then the constraints that we want to impose are 
    \[b^{\star} - \gamma \leq \frac{\sum_{i}y^{i}_M}{\sum_{i}y^{i}_M + \sum_{i}y^{i}_W} \leq b^{\star} + \gamma\]
    where $\gamma$ is an adjustable hyperparameter representing the error limit.
    These constraints can be generalized to $A\sum_i y^i - b \leq 0$ and a Lagrangian formulation can then be used to solve the problem.
    
    \subsection{Hate speech}
    Social media platforms are being criticized for being echo chambers of hate speech. However automated removal of such messages leads to suppressing the already marginalized voice of minorities. The task is especially challenging because what is considered toxic inherently depends on social context (e.g., speaker’s identity or dialect). Terms earlier deemed offensive (e.g., “n*gga”, “queer”) are no longer so if used by people from these groups \cite{sap-etal-2019-risk}. As race information is not available on Twitter, the authors use a proxy for identifying race, i.e the African American English dialect to signal race. The authors showed that the usage of the AAE dialect was shown to have high correlation with a tweet being classified as being hateful. Grouping tweets into White aligned and AAE aligned, the authors show that the false positive rates for AAE was 46\% as compared to 9\% for White aligned tweets.
    
    \subsection{Adversarial Learning Algorithms}
    Learning an generative adversarial network for the mitigating bias seems to be the most generalizable approach. \citet{10.1145/3278721.3278779} model a variation of GANs for mitigating biases. 
    
    In the context of learning word analogies, the generator learns the answers to simple word analogies while the discriminator tries to infer the gender from the prediction yielded by the generator. The generator attempts to prevent the discriminator from identifying the gender and will thus yield solutions which are less plagued by bias as they demonstrate in their paper. 
\end{enumerate}
\section{Why are current techniques not sufficient?}
\label{sec:current}
\citet{gonen-goldberg-2019-lipstick} point out that current methods of debiasing word vectors show promising scores on their seld-defined metrics. However, the removal is quite superficial. The actual effect is just hiding the bias, not removing it. Tests like WEAT and SEAT are good to check the presence of bias, if any, however small WEAT scores do not mean that bias has been removed. The gender bias is still reflected in the debiased embeddings, and can be recovered from any downstream model. They find that:-
\begin{enumerate}
    \item Words with strong previous gender bias (with the same direction) are easy to cluster together.
    \item The words with neutral gender, for eg receptionist, hairdresser which are sterotypically associated with the female gender still cluster together similar to the non-debiased word embeddings.
    \item A downstream classifier is still able to recover the implicit gender that was present in the debiased vectors just from their vectors and no additional information.
\end{enumerate}
Gender-direction which is the basis of gender association provides a way to measure the gender association of a word, but does not determine it. New metrics are needed in order to allow a metric to allow checking of bias in embeddings. Difference of accuracies over a gender-swapped GBET would be a good metric to check the bias but would not be a good method to train a network.

\section{Conclusion and Future Work}
\label{sec:future}
We looked at the different types of bias that exist in today's machine learning systems and how deploying them in the real world leads to harmful outcomes. We then tried to define fairness formally and showed how different definitions lead to different models. We showed the different metrics used to measure bias that may be present in word embeddings or models. We finally looked at a wide range of subfields of NLP where debiasing is done and the different methods through which this is realized. Finally, \citet{gonen-goldberg-2019-lipstick} pointed out that these metrics require a close look as they act as good indicators but do not determine it.

This field of bias mitigation in NLP has recently come up with the increase in the real word deployment of AI systems. The black box nature of NLP systems amplifies any problem that may exist due to bias, as the people using the system are not aware of the reasoning behind a prediction, they just know the final answer. This is a hot field with new innovations coming up everyday, however there is still a lot of scope for work in this space.
\begin{itemize}
    \item Most research has been restricted to measuring gender bias. However, there is little research in measuring bias due to race, religion or class, as these are more difficult to measure and may be less prevalent in typically used corpora. 
    \item Most systems assume that we know which kind of harmful bias exits in the corpora. However, this may not always be true, as sometimes, the sources of bias may not be obvious. For eg:- A resume filtering system may be biased against hiring people from a certain locality, which may not be obvious to those who use the system.
    \item Current metrics, like WEAT measure the gender association of a gender neutral word, but do not determine the bias it contains. That is, some systems may be reducing the scores relating to these metrics but the system may still be making biased predictions. New metrics which determine human bias are required.
    \item The Computer Science community, conducting such research, may not be well versed with the origins and the problems that bias can bring about in the real world. Thus more inter-disciplinary research with social scientists may lead to new insights. A hard-hitting truth is that this research itself does not come from culturally or socially diverse backgrounds.
    \item Work in hate-speech and machine translation has uncovered that speaker information is very important in order to remove biases from the system. However, most corpora strip away the original source, or the demographics of the person writing the article or the context in which the article was published. This origin information is invaluable and may lead to removal of some biases.
\end{itemize}

\bibliography{acl2020}
\bibliographystyle{acl_natbib}
\end{document}